\documentclass{article}

\usepackage{PRIMEarxiv}

\usepackage[utf8]{inputenc} 
\usepackage[T1]{fontenc}    
\usepackage{hyperref}       
\usepackage{url}            
\usepackage{booktabs}       
\usepackage{amsfonts}       
\usepackage{nicefrac}       
\usepackage{microtype}      
\usepackage{lipsum}
\usepackage{fancyhdr}       
\usepackage{graphicx}       
\graphicspath{{media/}}     

\usepackage{authblk}
\usepackage{booktabs}
\usepackage{xspace}
\newcommand*{\eg}{e.g.\@\xspace}
\newcommand*{\ie}{i.e.\@\xspace}
\newcommand*{\tempa}{\multicolumn{1}{r|}{}}

\newcommand*{\etal}{et al\@\xspace}
\usepackage{pifont}
\newcommand{\cmark}{\ding{51}}%
\newcommand{\xmark}{\ding{55}}%
\usepackage{color}
\usepackage{array}
\usepackage{amsmath}
\usepackage{threeparttable}

\usepackage[linesnumbered,ruled,vlined]{algorithm2e}
\SetKwInput{KwInput}{Input}                
\SetKwInput{KwOutput}{Output}              
\SetKwInput{KwInitial}{Initialization}
\SetKwFor{For}{for}{}{}

\title{Bypass Network for Semantics Driven Image Paragraph Captioning
}

\author[1]{Qi Zheng}
\author[2]{Chaoyue Wang}
\author[3]{Dadong Wang}
\affil[1]{University of Sydney, NSW 2008, Australia}
\affil[2]{JD Explore Academy}
\affil[3]{DATA61, CSIRO, NSW 2122, Australia}

\begin{document}
\maketitle


\begin{abstract}
Image paragraph captioning aims to describe a given image with a sequence of coherent sentences. Most existing methods model the coherence through the topic transition that dynamically infers a topic vector from preceding sentences. However, these methods still suffer from immediate or delayed repetitions in generated paragraphs because (i) the entanglement of syntax and semantics distracts the topic vector from attending pertinent visual regions; (ii) there are few constraints or rewards for learning long-range transitions. In this paper, we propose a bypass network that separately models semantics and linguistic syntax of preceding sentences. Specifically, the proposed model consists of two main modules, \ie a topic transition module and a sentence generation module. The former takes previous semantic vectors as queries and applies attention mechanism on regional features to acquire the next topic vector, which reduces immediate repetition by eliminating linguistics. The latter decodes the topic vector and the preceding syntax state to produce the following sentence. To further reduce delayed repetition in generated paragraphs, we devise a replacement-based reward for the REINFORCE training. Comprehensive experiments on the widely used benchmark demonstrate the superiority of the proposed model over the state of the art for coherence while maintaining high accuracy.
\end{abstract}

\section{Introduction}
\label{sec:intro}

Visual captioning explores machines' capability to comprehend the visual world by describing it with natural language. Traditional image captioning~\cite{kulkarni2013babytalk,xu2015show,lu2017knowing,yao2017boosting,wu2019recall,wu2020fine,yang2018multitask,lu2022data,tan2022acort} depicts an image with a single sentence that covers the whole scene or the most salient object. Dense image captioning~\cite{johnson2016densecap,yang2017dense,yin2019context,kim2019dense} describes all the detected salient regions without regard to the order of descriptions. Later, \cite{krause2017hierarchical} introduced image paragraph captioning that aims to give a coherent fine-grained paragraph.

The primary prerequisite of generating a coherent paragraph description for an image is to model coherence by its linguistic definition. Referring to linguistics~\cite{daniel2019speech,grosz1995centering}, the coherence exists when nearby sentences (a sentence-cluster) are ``about" someone or something, and the whole paragraph does not jump back or forth among multiple entities. It indicates that the coherence flowing through a paragraph can be modeled by topic transition among sentences. 
Most existing methods~\cite{liang2017recurrent,chatterjee2018diverse,wu2019densely,krause2017hierarchical} attempt to guide the topic transition by dynamically inferring a topic vector at the beginning of each sentence. For instance, conditioning on preceding sentences, \cite{liang2017recurrent} transfers attended visual features into a topic vector. In the work of \cite{wu2019densely}, the topic vector is predicted from the state of a hierarchical attention module, which integrates both regional features and preceding sentences.

Compared with directly transferring image captioning to paragraph captioning, these topic vector-based methods can generate more relative and diverse sentences owing to considering the information contained in the former sentence. However, directly using the former sentence as conditional input signals may only achieve a semblance of coherence. In other words, though maintaining linguistic consistency, the generated sentences fail to capture the variance in descriptive content. As a result, these methods still suffer from immediate or delayed repetition in synthesized paragraphs. In \textit{immediate repetition}, the subsequent sentence is a duplicate of the previous one, either by copying words or expressing the same stuff. As for \textit{delayed repetition}, the subsequent sentence may go back to earlier descriptive content regardless of the coherence within the generated partial paragraph. 

According to our observation, the topic transition that is directly controlled by preceding sentences fails to guarantee the desired coherence due to two reasons. On the one hand, such a transition does not distinguish the linguistic consistency from the content variance, which results in a compromise. Ideally, the topic transition should be driven by the development of descriptive content. However, in existing methods, the entanglement of syntax and semantics in preceding sentences distracts the topic vector from attending pertinent visual regions. On the other hand, most existing methods lack direct supervision to guide the long-range topic transition. The widely used maximum likelihood estimation promotes accuracy in word prediction but provides little feedback to sentence generation in a given context.

In this paper, we propose a bypass network to solve these issues. The proposed bypass structure separates semantics from the linguistic syntax of preceding sentences to produce a topic vector. The semantic stream drives topic transition, and the syntax is maintained by a bypass Part-Of-Speech (POS) stream. The decoder integrates these two elements to generate the following sentence. In this way, the topic vector can attend more accurately to pertinent visual regions and reduce immediate repetition. Given the simplicity of the proposed model, we provide a detailed deduction of the disentangling property in the appendix material. In addition, we devise a replacement-based reward for REINFORCE training. It enhances the quality of long-range transitions and alleviates the delayed repetition in generated paragraphs. Finally, we conduct comprehensive experiments to demonstrate the efficacy of the proposed model. The main contributions of our paper are listed below:

\begin{itemize}
    \item We propose a Bypass network to model topic transition in image paragraph captioning. It drives topic transition by maintaining a semantic stream and reduces immediate repetition.
    \item We further devise a replacement-based reward to alleviates delayed repetition, which takes effect during REINFORCE training.
    \item We compare paragraph-level and sentence-level captioning performance to assess coarse- and fine-grain coherence and accuracy, respectively.
\end{itemize}

\section{Related Work} \label{sec:re_w}
\subsection{Image Captioning and Dense Captioning}
The image captioning task aims to depict a given image using one sentence that covers the semantic meaning of the whole scene or describes a salient object in the scene. Recent deep learning-based approaches~\cite{donahue2015long,karpathy2015deep,anderson2018bottom,jiang2018recurrent,lu2017knowing,vinyals2015show,xu2015show} generally employ an encoder-decoder framework, which first extracts features from the image using CNNs~\cite{lecun1995convolutional} and then inputs them to a Recurrent Neural Network such as LSTMs~\cite{graves2013speech}. Visual features range from CNN features, grid features to object proposals. There are also some attempts that introduce object attributes or visual relationships to boost the semantic quality of the generated caption~\cite{yao2017boosting,yao2018exploring}. As for the decoder, attention mechanism~\cite{xu2015show,anderson2018bottom} has been introduced as a popular tool to attend and visualize where the model ``sees'' when it depicts.

Dense image captioning was introduced to enrich the descriptive details in an image. \cite{johnson2016densecap} propose a fully convolutional localization network (FCLN) that detects object regions and generates regional annotations in a single forward pass. To reduce redundancy in detected bounding boxes and increase semantic saliency of object regions, \cite{yang2017dense} incorporate joint inference and context fusion in the localization and captioning process, where the image-level feature served as the contextual cues. To capture more details during the captioning, \cite{yin2019context} further utilize the neighboring feature and the attribute of each object region.

\subsection{Image Paragraph Captioning}
Paragraph descriptions can be generated either as a ``big'' sentence or as a sequence of sentences. The former works in an image-captioning way, where the most significant problem is the internal repetition. Upon the well-known Up-Down~\cite{anderson2018bottom} model, \cite{melas2018training} suggest suppressing the repetition by blocking the trigrams that have appeared before, denoted by TDC (Training for Diversity in image paragraph Captioning). This simple yet efficient technique gains a vast improvement in generated sentences with respect to automatic evaluation metrics. TEB~\cite{gupta2021text} further generates a paragraph-vector using a Text Embedding Bank (TEB) to guide the captioning process.

We conclude that the success of TDC is twofold. First, since the decoder is trained to highlight words that co-occur frequently in the training corpus, suppression on the trigrams that have been generated gives way to the remaining highly correlated words. This way, TDC avoids producing repeated words while ensuring the quality of predicted words. Second, existing automatic evaluation metrics on image paragraph captioning take the generated description as one ``big'' sentence, where the order of sentences makes no difference. However, a primary concern is that sentence-level coherence is not guaranteed in the highly-scoring TDC.

Alternatively, many recent paragraph captioning methods learn to generate a sequence of sentences. For instance, \cite{krause2017hierarchical} introduce a hierarchical framework (Regions-Hierarchical), where the sentence-level RNNs produce $K$ topic vectors for $K$ sentences and predict the probability that each vector indicates the ending of the paragraph. Unlike Regions-Hierarchical, \cite{liang2017recurrent} propose a generator where the two-level RNNs work as a whole and adopt a sentence discriminator and a topic-transition discriminator to assess sentence plausibility and topic coherence. \cite{chatterjee2018diverse} produce all the topic vectors in advance similar to Regions-Hierarchical, but then combine the topic vector with the semantic of the preceding sentence as the input to the word-level decoder. \cite{mao2018show} introduce the LDA-Topic model to generate a topic distribution for each sentence, where the topic is represented by an average of several discrete abstract conceptual embeddings clustered from the training corpus. To increase the supervision of topic transition, \cite{wu2019densely} propose a densely supervised hierarchical policy-value (DHPV) network that dynamically assesses the contribution of the currently generated sentence or word. \cite{yan2020paracnn} use a hierarchical CNN architecture (\ie ParaCNN) to extract contextual information between sentences and generate a visual paragraph. Similarly, Li \etal~\cite{li2020dual} also use CNNs for paragraph captioning.

Though these approaches learn sentence-level topic transition, most of them ignore the gap between linguistic consistency and content transition and suffer from immediate or delayed repetition. Our method follows the branch of modeling topic transition among all generated sentences, but differently, we separate the semantics from the syntax of preceding sentences in predicting the topic vector, which reduces the immediate repetition. Furthermore, we devise a replacement-based reward to alleviate the delayed repetition.

\begin{figure*}
    \centering
    \includegraphics[width=.95\textwidth]{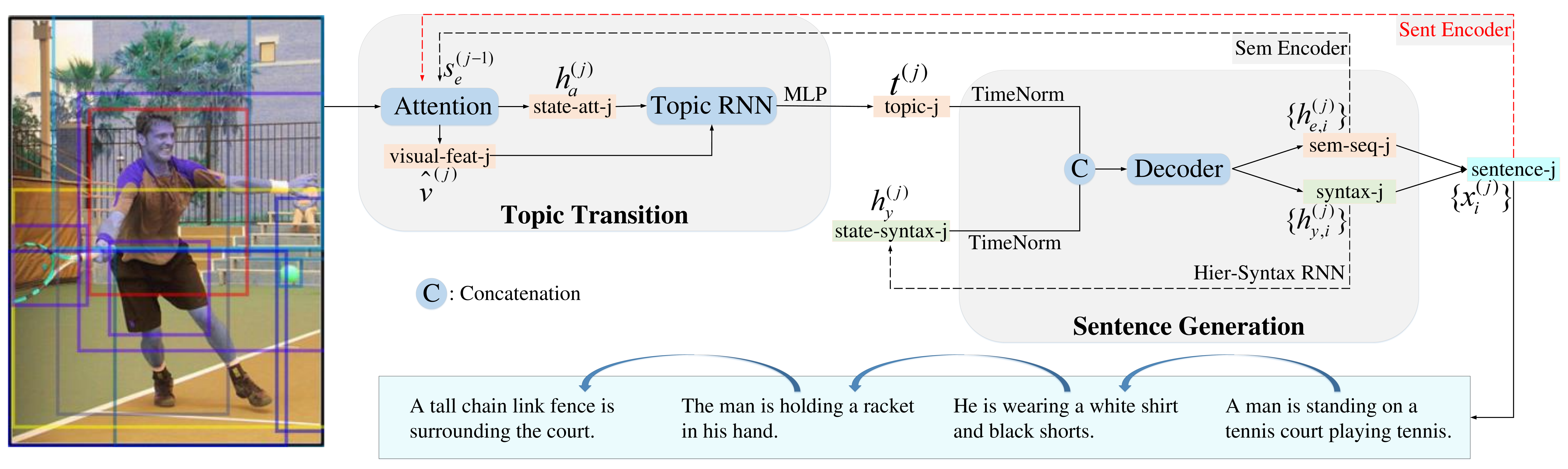}
    \caption{Illustration of generating the $j$-th sentence by our Bypass model. The model consists of i) the Topic-Transition module that predicts a topic vector based on preceding semantic stream, and ii) the Sentence Generation module that synthesizes a descriptive sentence given the topic vector and a syntactic state.}
    \label{fig:framework}
\end{figure*}

\section{Method}
To enhance topic transition, we design a bypass network that separates semantics from syntax in producing the topic vector and combines them in sentence generation. We first provide an overview of the proposed method and then explain the two parts of our model in detail, namely topic transition and sentence generation. Finally, we introduce the optimization process and discuss the replacement-based reward.

\subsection{Overview} \label{sec:overview}
Similar to~\cite{krause2017hierarchical,liang2017recurrent,melas2018training}, we first detect object regions in a given image $I$ using offline object detectors. This produces a set of regional features $\{v_1,v_2,\dots,v_K\}$, as the input to the pipeline in Fig.~\ref{fig:framework}. At the $j$-th time step, the Topic-Transition module first predicts an attention distribution $\{a_1,a_2,\dots,a_K\}^{(j)}$ over the regional features, conditioned on the semantics $s_e^{(j-1)}$ from the previous sentence. This gives the attended visual feature $\hat{v}^{(j)}$ and also the state $h_a^{(j)}$ of the attention network. 

The Topic RNN takes in these two elements as input and produces a topic hidden state $h_t^{(j)}$. The topic vector $t_j$ is generated by applying an MLP block on state $h_t^{(j)}$. The Sentence-Generation module applies the proposed Time-Normalization on the topic vector $t_j$ and the syntax state $h_y^{(j)}$ from preceding sentences, respectively. Then the results are concatenated and input into the decoder to produce a sequence of words. At each step, the hidden state from the decoder evolves into two streams, \ie the semantic stream and the syntactic stream. The semantic sequence is encoded by Semantic Encoder and forms the condition to the Topic-Transition module, while the syntactic stream is encoded by hierarchical-syntax (Hier-Syntax) Encoder and outputs the syntactic state of the $(j{+}1)$-th sentence.

\subsection{Topic Transition} \label{sec:topic_trans}
The Topic-Transition module determines the semantic content of the subsequent sentence, which can be seen as an extension of the Up-Down model~\cite{anderson2018bottom}. In traditional image captioning, the Up-Down model dynamically predicts attention distribution in generating each word. 

Differently, in image paragraph captioning, the dynamic is generally in sentence level, \ie all the words in one sentence share the same visual attention. Specifically, given the regional features $\{v_1,v_2,\dots,v_K\}$, the Attention RNN predicts the attention distribution for the $j$-th sentence,
\begin{align}
    h_a^{(j)} &= \mathrm{RNN}_a([s_e^{(j-1)};\bar{v}],h_a^{(j-1)}), \\
    a_k^{(j)} &= \mathrm{softmax}\left(w_a^T tanh(W_{va} v_k+W_{ha} h_a^{(j)})\right),
\end{align}
where $s_e^{(j-1)}$ is the semantic information of the $(j{-}1)$-th sentence and $\bar{v}=1/K\sum_k v_k$ is the mean-pooled regional feature. This gives the attended visual feature $\hat{v}^{(j)}=\sum a_k^{(j)}v_k$ for generating the $j$-th sentence.

Instead of directly being the topic vector as in the Up-Down model, $\hat{v}^{(j)}=\sum a_k^{(j)}v_k$ further passes a Topic RNN followed by an MLP block, and finally produces the topic vector $t^{(j)}$ and a probability $p^{(j)}$ of the $j$-th sentence being the last one,
\begin{align}
    h_t^{(j)}&=\mathrm{RNN}_t([\hat{v};h_a^{(j)}],h_t^{(j-1)}), \\
    t^{(j)}&=\mathrm{MLP}(h_t^{(j)})~~\mathrm{and}~~p^{(j)}=\mathrm{FC}(h_t^{(j)}).
\end{align}

The necessity of the Topic RNN comes from the fact that some descriptions in a paragraph may depend more on an abstract concept than the specific objects in a scene, which is beyond the coverage of the attended visual features. More details and the verification are provided in our experimental section. It is worth noting that this module is very close to the hierarchical policy network in~\cite{wu2019densely}, except that the information of the preceding sentence is from a semantic stream. We consider this a simple yet efficient baseline module and made minimum modifications to it.

\subsection{Sentence Generation} \label{sec:sent_gen}
To reduce internal covariate shift of the vectors from different time steps, inspired by the well-known Batch Normalization~\cite{ioffe2015batch}, we propose Time Normalization for the topic vector $t^{(j)}\in \mathbb{R}^D$ and the syntax state $h_y^{(j)}\in \mathbb{R}^D$, 
\begin{align}
    \hat{t}^{(j)} = \gamma_t \frac{t^{(j)}{-}\mu_t^{(j)}}{\sqrt{\sigma_t^{(j)}{+}\epsilon}}{+}\beta_t, \quad
    \hat{h_y}^{(j)} = \gamma_y \frac{h_y^{(j)}{-}\mu_y^{(j)}}{\sqrt{\sigma_y^{(j)}{+}\epsilon}}{+}\beta_y,
\end{align}
where $\gamma_\cdot,~\beta_\cdot \in \mathbb{R}^D$ are parameters to be learned. During multiple time steps, the mean of the topic vector is $\mu_t^{(j)}=1/j\sum_{m=0}^j t^{(m)}$ and the variance is $\sigma_t^{(j)}=1/j\sum_{m=0}^j (t^{(m)}-\mu_t^{(j)})^2$, which is similar for the syntax state. For efficiency, the mean and the variance are updated in an iterative way,
\begin{align}
    \mu_t^{(j)}&=\frac{1}{j}\left((j-1) \mu_{t}^{(j-1)} + t^{(j)}\right), \\
    \sigma_t^{(j)}&=\frac{j-1}{j}\left(\sigma_t^{(j-1)}+\frac{(\mu_t^{(j-1)}-t^{(j)})^2}{j}\right).
\end{align}

Given the normalized topic vector $\hat{t}^{(j)}$ and syntax state $\hat{h_y}^{(j)}$, the decoder produces a hidden state for each word,
Then the hidden state $h_{d,i}^{(j)}$ splits into two branches, \ie the semantic stream $h_{e,i}^{(j)}=\mathrm{MLP}(h_{d,i}^{(j)})$ and the syntax stream $h_{y,i}^{(j)}=\mathrm{MLP}(h_{d,i}^{(j)})$, and the word $x_i^{(j)}$ is predicted by
\begin{align} \label{eq:add}
    x_i^{(j)} = \mathrm{softmax}\left(\mathrm{FC}(h_{e,i}^{(j)}+h_{y,i}^{(j)})\right).
\end{align}
The sequence of semantic elements $h_{e,i}^{(j)}$ is encoded to form the semantic vector $s_e^{(j)}=\mathrm{RNN}_e(h_{e,i}^{(j)})$, as the $(j{+}1)$-th input to the Topic-Transition module. The sequence of syntax elements $h_{y,i}^{(j)}$ is decoded into POS tags $y_i^{(j)}$ of the $j$-th sentence, which then passes a sentence-level (Hier-Syntax) RNN to produce the unnormalized syntax state $h_y^{(j+1)}$.

\subsection{Optimization} \label{sec:optim}
\subsubsection{Training by cross-entropy loss.}
Given the available text $\mathbf{x}_*$ and POS tags $\mathbf{y}_*$, gram-level cross-entropy loss is applied to the generated word-sequence and tag-sequence, respectively,
\begin{align}
    \mathcal{\ell}_x{=}{-}\log(P(\mathbf{x}^{(j)}{=}\mathbf{x}_*^{(j)})){=}{-}\sum_i\log(P(x_i^{(j)}{=}x_{i,*}^{(j)}) \\
    \mathcal{\ell}_y{=}{-}\log(P(\mathbf{y}^{(j)}{=}\mathbf{y}_*^{(j)})){=}{-}\sum_i\log(P(y_i^{(j)}{=}y_{i,*}^{(j)})
\end{align}
A sentence-level cross-entropy loss is applied to the probability $p^{(j)}$ that indicates the ending sentence,
\begin{align}
    \mathcal{\ell}_p=-\sum_j\log(P(p^{(j)}=p_*^{(j)})),
\end{align}
where $p_*^{(j)}=1$ if the $j$-th sentence is the last one, otherwise $p_*^{(j)}=0$. Finally, the total loss is a weighted sum of the three items,
\begin{align} \label{eq:loss_xe}
    \mathcal{L}_\mathrm{xe}=\mathcal{\ell}_x+\lambda_y\mathcal{\ell}_y+\lambda_p\mathcal{\ell}_p,
\end{align}
where $\lambda_y$ and $\lambda_p$ are hyper-parameters. 

It is worth noting that we add no extra constraint on $h_{e,i}^{(j)}$ and $h_{y,i}^{(j)}$ to reduce their correlation. We proved that minimizing the loss $\mathcal{L}_\mathrm{xe}$ is equal to minimizing the off-diagonal entries of the function’s Hessian matrix \textit{w.r.t.} its inputs $\mathbf{x}$ and $\mathbf{y}$, which encourages the disentanglement between the them. Therefore, no extra constraint is compulsory to reduce the correlation of $h_{e,i}^{(j)}$ and $h_{y,i}^{(j)}$.

\subsubsection{Training by REINFORCE algorithm.}
From the viewpoint of Reinforcement Learning, topic transition in image paragraph captaining can be modeled as a Markov reward process, because the transition (from \textit{state} to \textit{action}) probability is deterministic. In \cite{wu2019densely}, an incremental sentence-level reward $r^{(j)}$ is proposed to assess the contribution of each sentence to the description paragraph,
\begin{align} \label{eq:reward_vanilla}
    r(\mathbf{y}^{(j)}) = \phi(\mathbf{y}^{(1:j)},G)-\phi(\mathbf{y}^{(1:j-1)},G),
\end{align}
where $\phi(\cdot)$ is a function that returns the value of the generated partial paragraph by comparing it with the annotated paragraph $G$.

For further training, an efficient solution is using the REINFORCE algorithm~\cite{williams1992simple} to maximize the expected total reward,
\begin{align} \label{eq:mrp}
    \mathcal{J}(\theta)=\sum_j\sum_{\mathbf{y}^{(j)}} p_\theta(\mathbf{y}^{(j)}|v_{1:K},\mathbf{y}^{(1:j-1)}) r(\mathbf{y}^{(j)}),
\end{align}
where $\theta$ is the set of parameters that have been defined. For simplicity, we denote $z=[v_{1:K},\mathbf{y}^{(1:j-1)}]$ and then derive
\begin{align} \label{eq:reinforce}
    \nabla_\theta\mathcal{J}(\theta)&=\sum_j\sum_{\mathbf{y}^{(j)}}r(\mathbf{y}^{(j)})\nabla_\theta\log p_\theta(\mathbf{y}^{(j)}|z) \nonumber \\
    & \approx \sum_j\frac{1}{N}\sum_{n=1}^N r(\mathbf{y}^{(j,n)})\nabla_\theta\log p_\theta(\mathbf{y}^{(j,n)}|z),
\end{align}
where $\mathbf{y}^{(j,n)},~1\leq n\leq N$ are the sentence sampled by Monte Carlo methods~\cite{kalos2009monte}.

\textbf{Replacement-based reward.}
One of the main drawbacks of the vanilla algorithm in Eq.~(\ref{eq:reinforce}) is its high variance. In the well-known SCST~\cite{rennie2017self} algorithm for single-sentence captioning, a greedy-search baseline is introduced to reduce the variance. We could similarly set a baseline for the reward in Eq.~(\ref{eq:reward_vanilla}) as $r(y^{(j)})-r_b(y^{(j)})$,
\begin{align}
r_b(y^{(j)})=\phi([\mathbf{y}^{(1:j-1)};\hat{\mathbf{y}}^{(j)}], G)-\phi(\mathbf{y}^{(1:j-1)}, G),
\end{align}
where $\hat{\mathbf{y}}^{(j)}$ is sampled by greedy search from the state policy $p_\theta(\mathbf{y}^{(j)}|z)$. However, we observed that such high variance comes much more from the reward itself. Specifically, an obvious difference of the length (\ie numbers of words) between the generated partial paragraph $\mathbf{y}^{(1:j)}$ and the annotated paragraph $G$ seriously undermines the estimation of the real contribution of $\mathbf{y}^{(j)}$.

On the other hand, Eq.~(\ref{eq:reward_vanilla}) only counts the topic transition between adjacent sentences, \ie $\mathbf{y}^{(j)}$ and $\mathbf{y}^{(j-1)}$, which is effective for immediate repetition but not the delayed repetition. Therefore, we devise a replacement-based reward to address both issues,
\begin{align}
    r(\mathbf{y}^{(j)}) = \phi(\mathbf{y}^{(1:j)},\mathbf{y}_*^{(1:j)})-\phi(\tilde{\mathbf{y}}^{(1:j)},\mathbf{y}_*^{(1:j)}),
\end{align}
where $\tilde{\mathbf{y}}^{(1:j)}=[\mathbf{y}^{(1:j-1)};\tilde{\mathbf{y}}^{j}]$ is the mixed partial description and $\phi(\tilde{\mathbf{y}}^{(1:j)},\mathbf{y}_*^{(1:j)})$ serves as the baseline reward. $\tilde{\mathbf{y}}^{j}=\mathrm{sample}(\mathbf{y}^{1:j-1})$ is the $j$-th sentence randomly sampled from previously generated $j{-}1$ sentences. This can also be optimized by Eq.~(\ref{eq:reinforce}).

\section{Experiments}
We conduct experiments on the Stanford image-paragraph dataset\footnote{\url{https://cs.stanford.edu/people/ranjaykrishna/im2p/index.html}}~\cite{krause2017hierarchical}. This dataset contains $19,551$ images from MS COCO dataset~\cite{lin2014microsoft} and Visual Genome~\cite{krishna2017visual}, each of which is labeled with one paragraph of about $67.50$ words. Each paragraph consists of multiple sentences with $11.91$ words/sentence on average. It has been officially divided into training ($14,575$), validation ($2,487$) and test ($2,489$) splits.

\subsection{Implementation Details}
Following \cite{anderson2018bottom}, we use the Faster R-CNN~\cite{ren2015faster} object detector to obtain $K=36$ object proposals. The dimension of extracted regional features is $2,048$. Unless otherwise specified, all the RNNs are one-layer GRU blocks, with $512$ hidden units. Similar to~\cite{krause2017hierarchical,liang2017recurrent,wu2019densely}, we set the number of sentences to $6$ and the maximum words in a sentence to $30$. All the MLP blocks consist of two linear layers, connected by one ReLU activation layer and one dropout layer. The dropout probability during training is $0.5$. The dimension of word embeddings is $512$, which are trained from scratch. We replace the words that appear less than four times in the dataset with `unk' token and build a vocabulary of $4636$ words. POS tags of the annotated paragraphs are extracted by NLTK tools\footnote{\url{http://www.nltk.org/book/ch05.html}} and then merged into $17$ categories. The embedding size of POS tags is set as $64$. 

We train the proposed model to minimize Eq.~(\ref{eq:loss_xe}) using the Adam~\cite{kingma2014adam} optimizer with an initial learning rate of $0.0005$ for $70$ epochs. We set the weights of $\mathcal{\ell}_x$, $\mathcal{\ell}_y$ and $\mathcal{\ell}_p$ as $5{:}5{:}1$. The batch size throughout training is set to $10$. The learning rate decays every three epochs, with a decay rate of $0.8$. We further train the model to maximize Eq.~(\ref{eq:mrp}) for $30$ epochs with the same decay rate. One sample is used for the approximation in Eq.~(\ref{eq:reinforce}). We fix the beam-search size at $1$ for our model throughout the following experiments.

\subsection{Comparison Methods and Metrics}
We compare our model with previous approaches reported in~\cite{krause2017hierarchical}, including Sentence-Concat, Template, DenseCap-Concat and Image-Flat~\cite{karpathy2015deep} and the latest state-of-the-art methods, including Regions-Hier~\cite{krause2017hierarchical}, Up-Down~\cite{anderson2018bottom}, RTT-GAN~\cite{liang2017recurrent}, TDC~\cite{melas2018training}, 
ParaCNN~\cite{yan2020paracnn} and TEB~\cite{gupta2021text}. 
Since the pre-trained models are not provided, we followed the officially released codes for training TDC\footnote{\url{https://github.com/lukemelas/image-paragraph-captioning}}~\cite{melas2018training}, ParaCNN\footnote{\url{https://github.com/Shiyang-Yan/ParaCNN}}~\cite{yan2020paracnn}, and TEB\footnote{\url{https://github.com/arjung128/image-paragraph-captioning}}~\cite{gupta2021text}.
Codes for the remaining works are unavailable, so we implemented the basic form of DHPV as our baseline model. The implementation follows the original paper~\cite{wu2019densely} and training is under the same setting for a fair comparison.

\setlength{\tabcolsep}{8pt}
\begin{table*}[!t]
    \centering
    \begin{tabular}{l|c|c|c|c|c|c}
        \toprule
         & METEOR & CIDEr$^*$ & Bleu@4 & Bleu@3 & Bleu@2 & Bleu@1 \\ \hline
        TDC~\cite{melas2018training} & $8.81\pm 0.17$& $72.20\pm 1.89$ & $3.82\pm 0.13$ & $6.47\pm 0.18$ & $11.17\pm 0.25$ & $21.39\pm 0.41$ \\
        TEB~\cite{gupta2021text}& $8.10\pm 0.17$ & $66.10\pm 1.75$ & $3.63\pm 0.13$ & $6.07\pm 0.17$ & $10.50\pm 0.24$ & $19.84\pm 0.31$  \\
        ParaCNN~\cite{yan2020paracnn}& $8.52\pm 0.16$ & $61.06\pm 1.54$ & $3.89\pm 0.11$ & $6.87\pm 0.16$ & $12.11\pm 0.22$ & $23.11\pm 0.29$  \\
        Bypass & $\textcolor{red}{10.48\pm 0.16}$ & $\textcolor{red}{73.07\pm 1.48}$ & $\textcolor{red}{3.94\pm 0.11}$ & $\textcolor{red}{7.09\pm 0.16}$ & $\textcolor{red}{12.65\pm 0.23}$ & $\textcolor{red}{24.44\pm 0.29}$  \\
        \bottomrule
    \end{tabular}
    \caption{Sentence-level scores of different methods. The margin of error (\ie $\pm \epsilon$) is calculated at confidence interval $95\%$.}
    \label{tab:sent}
\end{table*}

\setlength{\tabcolsep}{2pt}
\begin{figure*}[!ht]
    \centering
    \begin{tabular}{cccc}
    \includegraphics[width=0.24\textwidth]{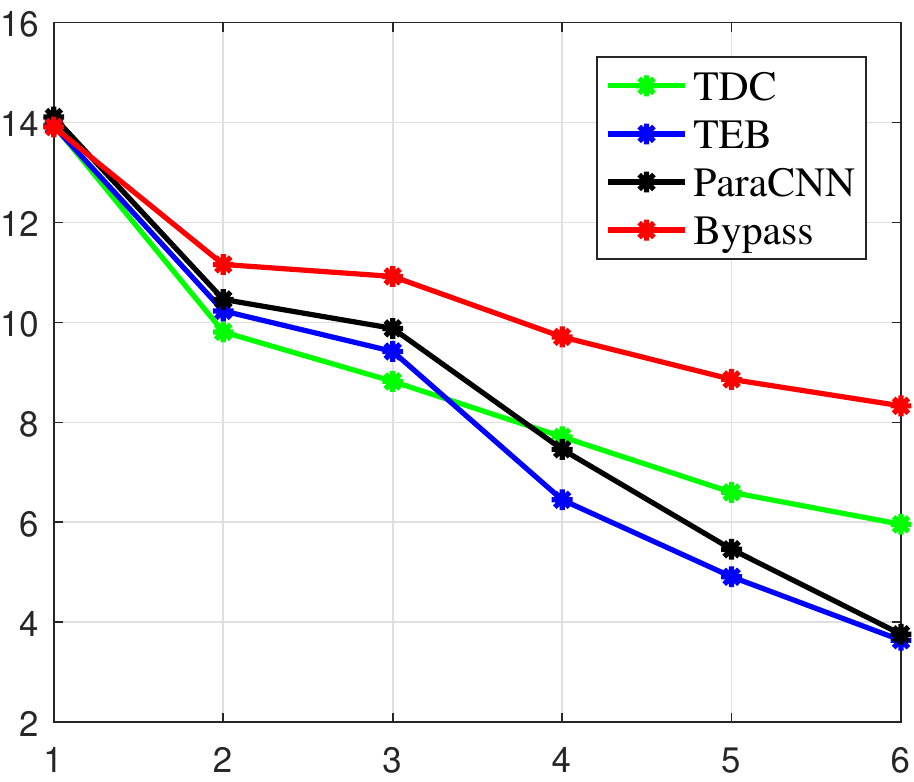} & 
    \includegraphics[width=0.24\textwidth]{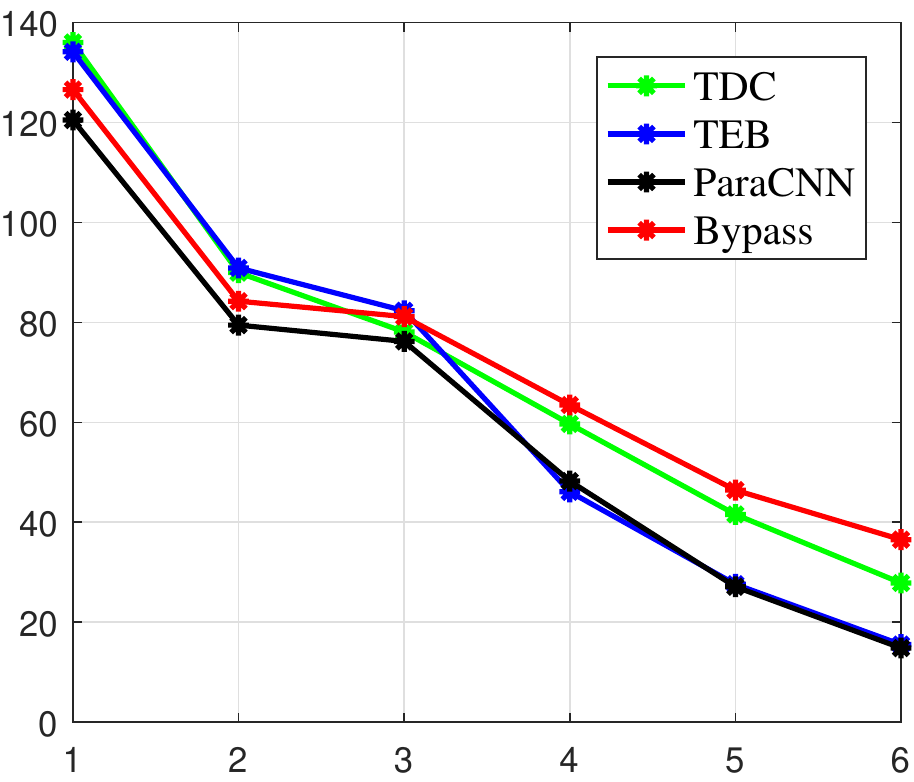} & 
    \includegraphics[width=0.24\textwidth]{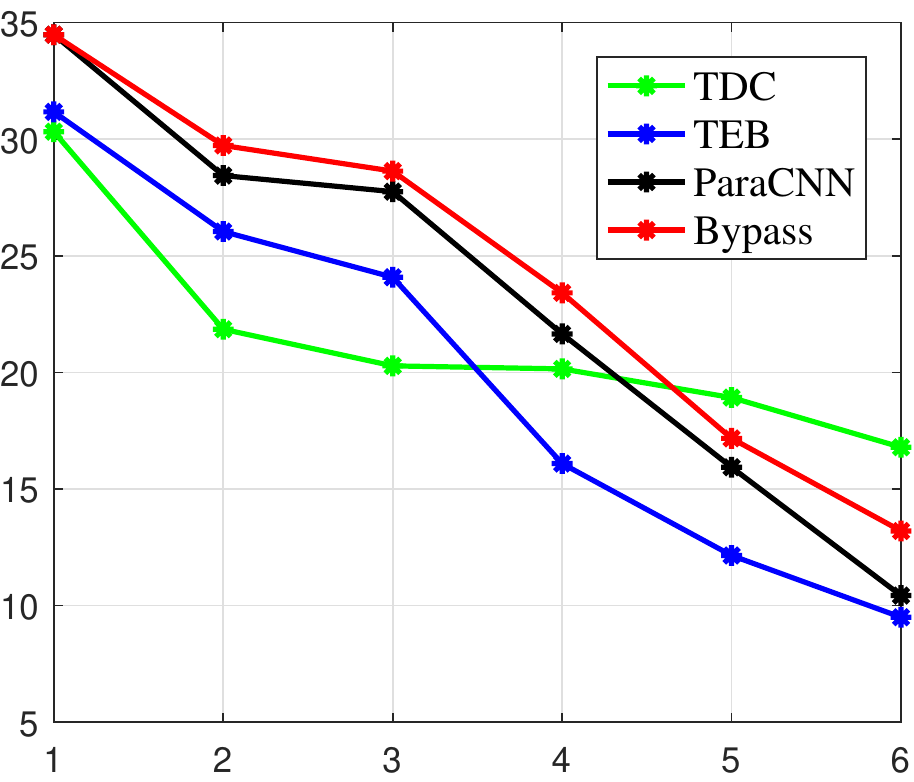} & 
    \includegraphics[width=0.24\textwidth]{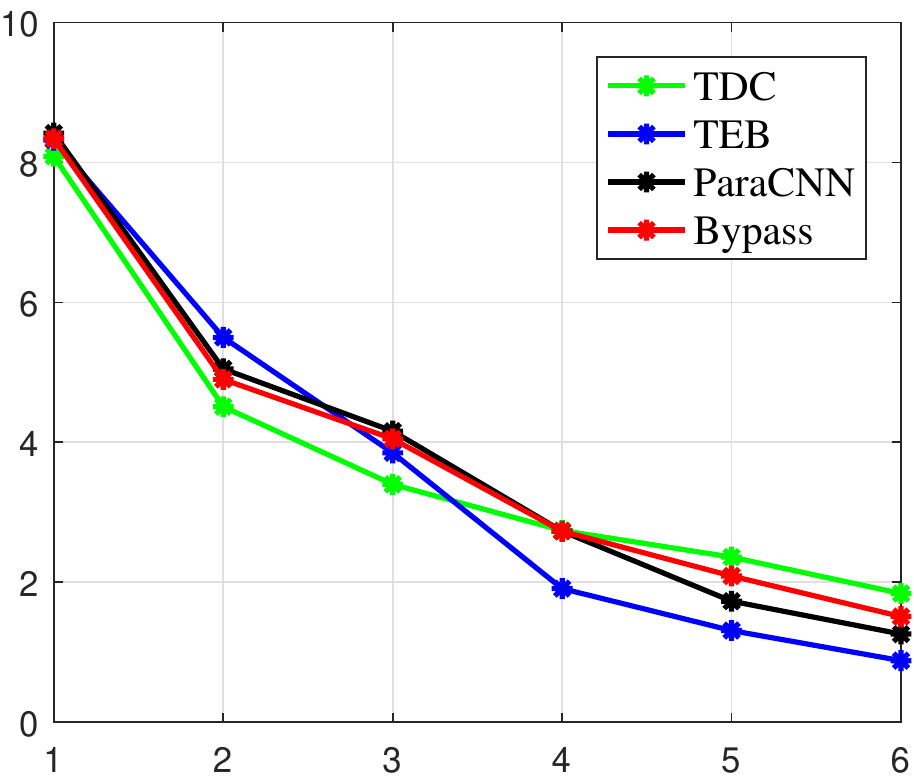} \\
    (a) METEOR & (b) CIDEr$^*$ & (c) Bleu@1 & (d) Bleu@4
    \end{tabular}
    \caption{Sentence-level score curves of different methods. The index of the x-axis denotes the first to the sixth sentence.}
    \label{fig:metrics}
\end{figure*}

We report six automatic evaluation metrics, \ie Bleu@1-4~\cite{papineni2002bleu}, METEOR~\cite{banerjee2005meteor} and CIDEr~\cite{vedantam2015cider}. Among them, Bleu@1-4 scores measure the precision of n-gram matching. METEOR assesses the unigram matching based on the surface form, stemmed form, and meaning of the unigrams. CIDEr evaluates the consensus of generated captions based on TF-IDF weighting for each n-gram. First, following existing sentence/dense captioning methods, we treat the generated paragraph as a big sentence to compare these metrics. However, such measurement is insufficient to reflect the coherence within the generated paragraph, since it ignores the order of sentences within a paragraph. Therefore, we further measure sentence-level scores of these six metrics. Specifically, each sentence in a generated paragraph is compared with the corresponding sentence of the annotated paragraph, and the average over six sentences gives the final (Bleu@n, METEOR or CIDEr) score of a sample. Note that, since the distribution error of document frequency in the single-sentence ground truth, we replace CIDEr with CIDEr$^*$ that removes the IDF term. 

\subsection{Comparisons with Previous Methods}

\paragraph{Paragraph-level evaluation}
As shown in Table~\ref{tab:res}, the methods Sentence-Concat, Template, DenseCap-Concat, Image-Flat~\cite{karpathy2015deep} and Up-Down~\cite{anderson2018bottom} perform poorly \textit{w.r.t} both the n-gram metrics and the census-based metric. We examined the generated paragraphs from the Up-Down model and found that the description sentences in a paragraph are heavily repeated. 
Under cross-entropy training, Regions-Hier~\cite{krause2017hierarchical} achieves the highest Bleu@1 and METEOR scores, but a low CIDEr score. This reveals that the model learns overall accuracy but fails to predict those semantic words. On the contrary, TDC gains a high CIDEr score and a low Bleu@1 score. Our model integrates the advantages of the ``big'' sentence methods and the topic-based methods, and presents relatively stable paragraph-level coherence and accuracy. With adversarial training and extra data, RTT-GAN performs best \textit{w.r.t} METEOR and Bleu scores, followed by our model. However, our model surpasses RTT-GAN by $27.90\%$ on the CIDEr score, which validates the effectiveness of the replacement-based reward.

\setlength{\tabcolsep}{6pt}
\begin{table}[!th]
    \centering
    \begin{tabular}{l|c|c|c|c|c}
        \toprule
         & METER & CIDEr & Bleu@1 & Bleu@4 & RL \\ \hline
        Sentence-Concat & 12.05 & 6.82 & 31.11 & 3.98 & \xmark \\
        Template & 14.31 & 12.15 & 37.47 & 7.38 & \xmark \\
        DenseCap-Concat & 12.66 & 12.51 & 33.18 & 4.54 & \xmark  \\ 
        Image-Flat & 12.82 & 11.06 & 34.04 & 7.71 & \xmark \\
        Up-Down & 13.66 & 12.89 & 32.78 & 6.89 & \xmark \\
        Regions-Hier$^*$ & \textcolor{red}{15.95} & 13.52 & \textcolor{red}{41.90} & \textcolor{blue}{8.69} & \xmark \\
        TDC & 15.63 & \textcolor{red}{23.96} & 37.29 & \textcolor{red}{8.98} & \xmark \\
        Bypass & \textcolor{blue}{15.77} & \textcolor{blue}{18.35} & \textcolor{blue}{40.23} & 8.65 & \xmark \\ \hline
        Up-Down & 13.63 & 13.77 & 29.67 & 5.88 & \cmark \\
        RTT-GAN$^*$ & \textcolor{red}{18.39} & 20.36 & \textcolor{red}{42.06} & \textcolor{red}{9.21} & \cmark \\
        TEB & 15.11 & 22.13 & 35.90 & 8.44 & \cmark \\
        ParaCNN & 14.96 & 18.77 & 37.37 & 7.69 & \cmark\\
        Bypass & \textcolor{blue}{17.27} & \textcolor{red}{26.04} & \textcolor{blue}{40.75} & \textcolor{blue}{9.05} & \cmark \\
        \bottomrule
    \end{tabular}
    \caption{Paragraph-level scores of different methods on the test split of the Stanford image-paragraph dataset. \xmark ~and \cmark represent training by cross-entropy loss and finetuning by REINFORCE algorithm, respectively. Among the \cmark, RTT-GAN and ParaCNN use adversarial training instead of RL. $^*$ indicates that extra data is used in training.}
    \label{tab:res}
\end{table}

\paragraph{Sentence-level evaluation} Results are listed in Table~\ref{tab:sent}. Compared with the ``big'' sentence model TDC that suppresses immediate repetition, the TEB module contributes noting to sentence accuracy and coherency. The ParaCNN achieves higher Bleu scores than TDC while lower METEOR and CIDEr$^*$ scores, which indicates increased accuracy and undermined semantic coherence. Our model surpasses TDC by $18.96\%$ METEOR and $1.23\%$ CIDEr$^*$ scores, and outperforms ParaCNN by $3.14\%$ Bleu@4 and $5.76\%$ Bleu@1 scores. It indicates that our model improves n-gram accuracy and enhances sentence coherence. This is further verified in Fig.~\ref{fig:metrics}. The proposed Bypass model consistently performs much better than the others \textit{w.r.t} METEOR on all the six sentences, slightly better than TDC \textit{w.r.t} CIDEr$^*$ and slightly better than ParaCNN \textit{w.r.t} Bleu scores. The improvements contribute to simultaneously overcoming immediate repetition and delayed repetition. 

\setlength{\tabcolsep}{5pt}
\begin{table}[!ht]
    \centering
    \begin{tabular}{l|c|c|c|c}
        \toprule
         & METER & CIDEr & Bleu@1 & Bleu@4  \\
        \hline
        Baseline & 15.48 & 16.40 & 38.83 & 8.34 \\
        Baseline + TimeNorm & 15.39 & 17.01 & 38.73 & 8.30 \\
        Bypass & \textcolor{red}{15.77} & \textcolor{red}{18.35} & \textcolor{red}{40.23} & \textcolor{red}{8.65} \\ \hline
        Baseline + R$_{\mathrm{increment}}$ & 17.36 & 18.51 & 39.86 & 8.46 \\
        Bypass + R$_{\mathrm{increment}}$ & 17.11 & 20.74 & \textcolor{red}{41.14} & \textcolor{red}{9.10} \\
        Bypass + R$_{\mathrm{replacement}}$ & \textcolor{red}{17.27} & \textcolor{red}{26.04} & 40.75 & 9.05 \\
        \bottomrule
    \end{tabular}
    \caption{Paragraph-level scores \textit{w.r.t} METEOR, CIDEr, Bleu@1 and Bleu@4.}
    \label{tab:abl_time_reward}
\end{table}

\subsection{Ablation Studies}

\paragraph{Paragraph-level evaluation}
As shown in Table~\ref{tab:abl_time_reward}, equipped with Time Normalization, our model surpasses the original baseline model on all metrics. We observed faster convergence in experiments and improved diversity in generated descriptions. We attribute the accelerated convergence to the reduced internal shift, where Time Normalization reduces the difference among the distributions of semantic vector or syntax vector at different time steps. As for diversity, the normalized variance avoids mode-collapse in the input vectors for each sentence. Trained by R$_{\mathrm{increment}}$, our model achieves higher scores than the baseline model \textit{w.r.t} all the metrics. Comparing our model trained with the two rewards, we can see a boost on the CIDEr metric when using the replacement-based reward. 

\setlength{\tabcolsep}{5pt}
\begin{table*}[!ht]
    \centering
    \begin{tabular}{l|c|c|c|c|c|c}
        \toprule
         & METEOR & CIDEr$^*$ & Bleu@4 & Bleu@3 & Bleu@2 & Bleu@1 \\ \hline
        Baseline & $9.11\pm 0.17$ & $71.39\pm 1.78$ & $\textcolor{red}{3.87\pm 0.13}$ & $6.79\pm 0.18$ & $11.99\pm 0.25$ & $22.79\pm 0.32$ \\
        Baseline+TimeNorm & $8.95\pm 0.16$ & $70.53\pm 1.70$ & $3.74\pm 0.12$ & $6.62\pm 0.17$ & $11.80\pm 0.24$ & $22.71\pm 0.32$  \\
        Bypass & $\textcolor{red}{9.32\pm 0.16}$ & $\textcolor{red}{71.93\pm 1.68}$ & $3.84\pm 0.11$ & $\textcolor{red}{6.95\pm 0.17}$ & $\textcolor{red}{12.30\pm 0.24}$ & $\textcolor{red}{23.51\pm 0.32}$\\ \hline
        Baseline+R$_{\mathrm{increment}}$ & $9.91\pm 0.15$ & $64.80\pm 1.35$ & $3.36\pm 0.09$ & $6.31\pm 0.15$ & $11.68\pm 0.21$ & $23.23\pm 0.27$ \\
        Bypass+R$_{\mathrm{increment}}$ & $10.15\pm 0.16$ & $71.64\pm 1.57$ & $\textcolor{red}{3.99\pm 0.11}$ & $\textcolor{red}{7.20\pm 0.17}$ & $\textcolor{red}{12.76\pm 0.24}$ & $24.24\pm 0.30$ \\
        Bypass+R$_{\mathrm{replacement}}$ & $\textcolor{red}{10.48\pm 0.16}$ & $\textcolor{red}{73.07\pm 1.48}$ & $3.94\pm 0.11$ & $7.09\pm 0.16$ & $12.65\pm 0.23$ & $\textcolor{red}{24.44\pm 0.29}$  \\
        \bottomrule
    \end{tabular}
    \caption{Sentence-level scores of different methods. The margin of error (\ie $\pm \epsilon$) is calculated at confidence interval $95\%$.}
    \label{tab:sent_abl}
\end{table*}

\begin{figure*}[!ht]
    \centering
    \includegraphics[width=.88\textwidth]{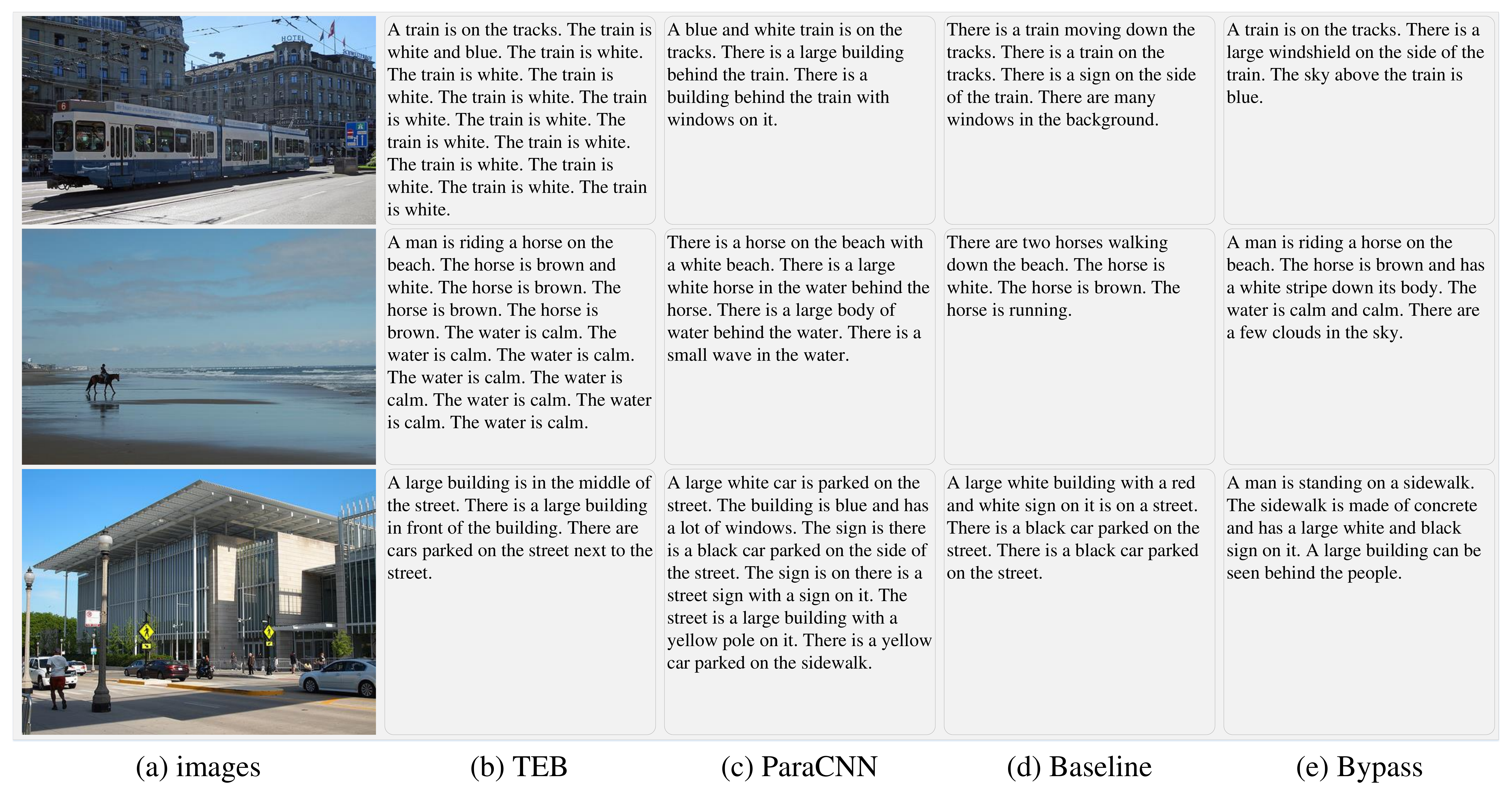}
    \vspace{-0.2cm}
    \caption{Descriptive captioning paragraphs generated by TEB, ParaCNN, the Baseline, and the Bypass model. 
    }
    \label{fig:quality}
\end{figure*}

\paragraph{Sentence-level evaluation} Table~\ref{tab:sent_abl} lists the sentence-level scores under cross-entropy training and REINFORCE training. Interestingly, Time Normalization undermines the performance of the Baseline. It verifies our hypothesis that linguistic information harms topic transition. Time Normalization cannot function effectively on the integration of linguistics and semantics. However, the improvement is obvious after these two aspects are distinguished by our Bypass model. From the results of REINFORCE training, the replacement-based reward enhances the sentence-level coherence while the increment-based reward promotes the sentence-level accuracy, complementary to the Bypass architecture.

\subsection{Qualitative Results} 
We list the descriptions generated by TEB, ParaCNN, the Baseline, and our model in Fig.~\ref{fig:quality}. For the first image, the paragraph from TEB contains heavy immediate repetition, which keeps describing ``the train is white''. ParaCNN improves the coherence of the paragraph by explicitly modeling topic transition, \ie from \textit{train}, \textit{building} to \textit{windows}. The Baseline model describes the visual scene fluently but has immediate repetition in the second sentence. Our model depicts around the \textit{train}, from \textit{track}, \textit{windshield} to \textit{sky}. As for the second image, similar repetition is observed in the captions from TEB, which is partially reduced by ParaCNN. The description from the Baseline is around the \textit{horse}, but not accurate enough and somewhat redundant. Our model seamlessly transits from \textit{man} and \textit{horse} to \textit{water} and \textit{sky}, with little repeated words. The third image is one of the very few samples that TEB depicts with little repetition. However, ParaCNN suffers from delayed repetition in this case. The same topic goes back and forth in different sentences, breaking the semantic coherence in the paragraph, \eg both the first and the last sentences are about \textit{car}. The caption generated by the Baseline is accurate yet contains immediate repetition in the last sentence. Our model resembles the accuracy of the Baseline and avoids repetition.

\begin{figure}[!h]
    \centering
    \includegraphics[width=.48\textwidth]{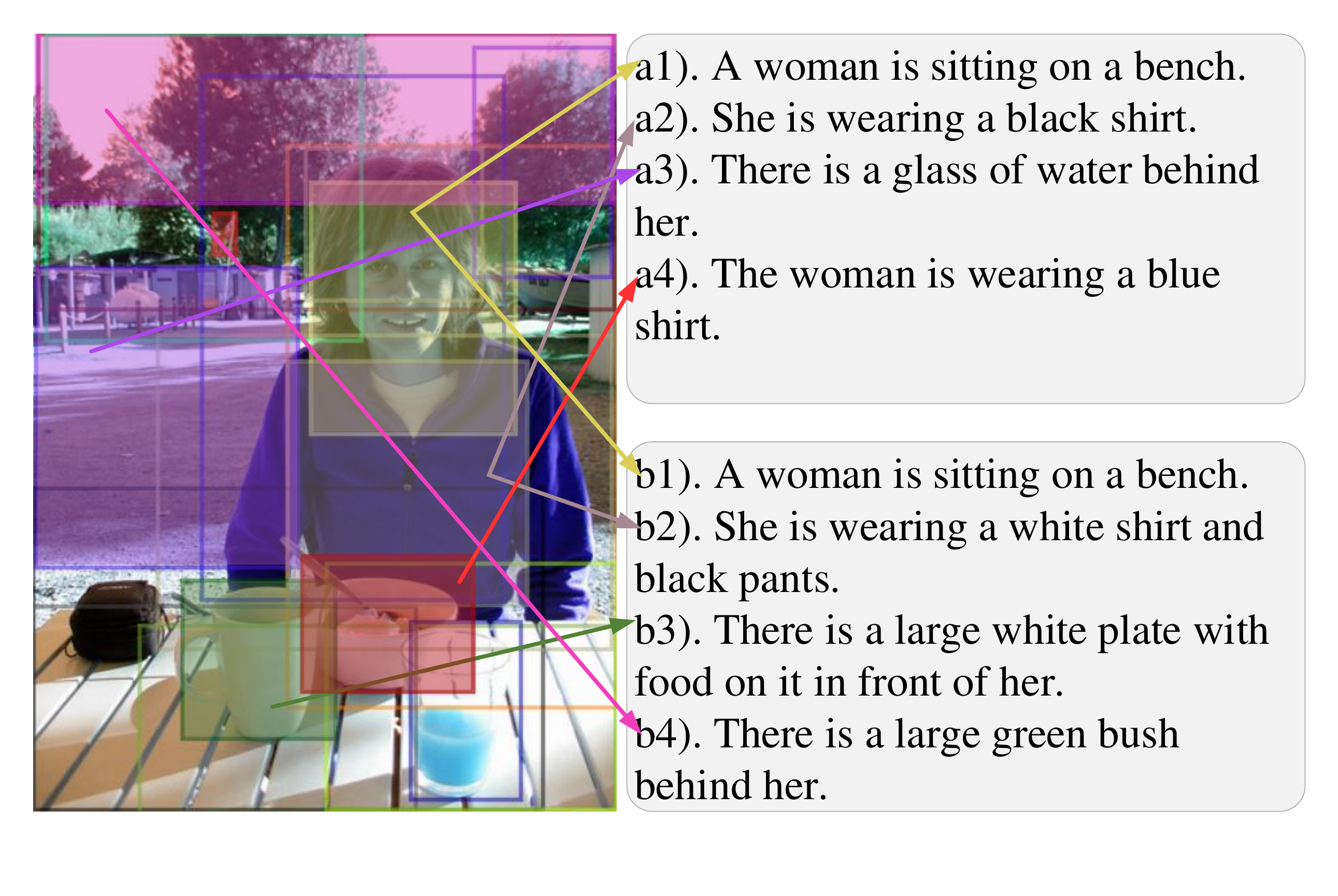}
    \caption{Visual attention during paragraph generation by the Baseline model (a1$\sim$a4) and the Bypass model (b1$\sim$b4).}
    \label{fig:att}
\end{figure}

\paragraph{Visualization of topic transition} As shown in Fig.~\ref{fig:att}, top visual areas are masked among the bounding boxes for each sentence in the paragraph captions by the Baseline (right up) and by our model (right down). 

We can see that both models capture accurate regions when producing the first two sentences, \ie, the \textit{woman} and the \textit{shirt}. However, for the third sentence, the Baseline model focuses on a \textit{street} area and depicts the item on the table. It continues describing \textit{woman} and \textit{shirt} in the fourth sentence while attending to the \textit{bowl}. Instead, our model first focuses and describes the \textit{cup} on the table in front of the woman in the third sentence, and then turns to the \textit{bush} behind her. The visual attention in our model moves more smoothly and has better control of the generated descriptions.

\begin{figure}[!h]
    \centering
    \includegraphics[width=.48\textwidth]{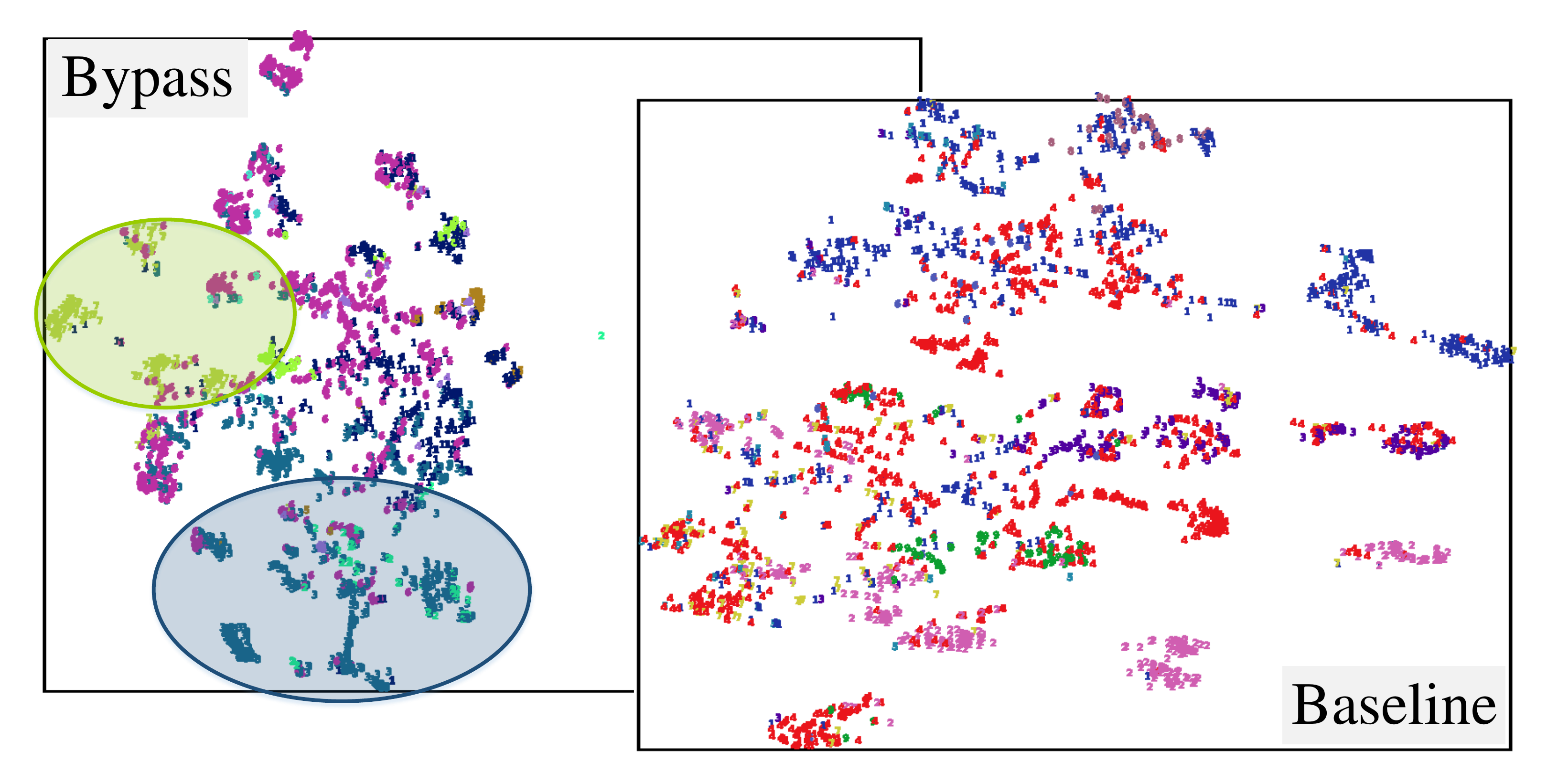}
    \caption{Clustering of topic vectors. Samples in the same color are from the same semantic category. The overlapped area covers very few points in the Bypass clustering space.}
    \label{fig:tsne}
\end{figure}

\paragraph{Visualization of topic clustering} Fig.~\ref{fig:tsne} shows the clustering space of the learned topic vectors from the Baseline and the proposed model. We can observe that the topic vectors of the Bypass model present a much better clustering effect \textit{w.r.t} the semantic. On the contrary, the topic vectors from the same semantic category spread more widely, and those from different categories mix with each other.

We circle two sets of clustered samples in space given by the Bypass model for illustration. The dark blue circle represents an outdoor scene, where the semantic words include \{`water', `ocean', `dirt', `trees', `tarmac', `snow', `rock', `sky', `grass', `road', `sand', `leaves', `gravel', `forest', `wooded', `mountain', `fence', `tracks', `ground', `sun', `enclosure', `hill', `slope', `boulder'\}. The light cyan represents an indoor scene, where the semantic words include \{`bottle', `luggage', `boxes', `attached', `window', `meter', `room', `handles', `bowls', `glass', `desk', `box', `door', `toilet', `plates', `suitcases', `shelf', `suitcase', `vase', `hanging', `bag', `sink', `bathroom', `kitchen'\}.

\section{Conclusion}
In this paper, we proposed a Bypass architecture that separates the semantic stream from sentences for a more coherent topic transition without extra disentangling loss. Moreover, a novel replacement-based reward was devised to reduce delayed repetition and the variance of the model when trained by the REINFORCE algorithm. Experiments showed improved sentence-level and paragraph-level coherence and accuracy of the proposed method without extra training data, and verified the validity of different modules. In the future, we would like to explore more on the syntactic stream to increase the linguistic diversity in generated paragraph descriptions.


\bibliographystyle{plain}
\bibliography{references}

\begin{thebibliography}{10}

\bibitem{anderson2018bottom}
Peter Anderson, Xiaodong He, Chris Buehler, Damien Teney, Mark Johnson, Stephen
  Gould, and Lei Zhang.
\newblock Bottom-up and top-down attention for image captioning and visual
  question answering.
\newblock In {\em CVPR}, pages 6077--6086, 2018.

\bibitem{banerjee2005meteor}
Satanjeev Banerjee and Alon Lavie.
\newblock Meteor: An automatic metric for mt evaluation with improved
  correlation with human judgments.
\newblock In {\em ACL Workshop}, pages 65--72, 2005.

\bibitem{chatterjee2018diverse}
Moitreya Chatterjee and Alexander~G Schwing.
\newblock Diverse and coherent paragraph generation from images.
\newblock In {\em ECCV}, pages 729--744, 2018.

\bibitem{daniel2019speech}
Jurafsky Daniel and Martin James~H.
\newblock Speech and language processing.
\newblock {\em Draft}, 2019.

\bibitem{donahue2015long}
Jeffrey Donahue, Lisa Anne~Hendricks, Sergio Guadarrama, Marcus Rohrbach,
  Subhashini Venugopalan, Kate Saenko, and Trevor Darrell.
\newblock Long-term recurrent convolutional networks for visual recognition and
  description.
\newblock In {\em CVPR}, pages 2625--2634, 2015.

\bibitem{graves2013speech}
Alex Graves, Abdel-rahman Mohamed, and Geoffrey Hinton.
\newblock Speech recognition with deep recurrent neural networks.
\newblock In {\em ICASSP}, pages 6645--6649. IEEE, 2013.

\bibitem{grosz1995centering}
Barbara~J Grosz, Scott Weinstein, and Aravind~K Joshi.
\newblock Centering: A framework for modeling the local coherence of discourse.
\newblock {\em Computational linguistics}, 21(2):203--225, 1995.

\bibitem{gupta2021text}
Arjun Gupta, Zengming Shen, and Thomas Huang.
\newblock Text embedding bank for detailed image paragraph captioning.
\newblock In {\em AAAI}, volume~35, pages 15791--15792, 2021.

\bibitem{ioffe2015batch}
Sergey Ioffe and Christian Szegedy.
\newblock Batch normalization: Accelerating deep network training by reducing
  internal covariate shift.
\newblock In {\em International conference on machine learning}, pages
  448--456. PMLR, 2015.

\bibitem{jiang2018recurrent}
Wenhao Jiang, Lin Ma, Yu-Gang Jiang, Wei Liu, and Tong Zhang.
\newblock Recurrent fusion network for image captioning.
\newblock In {\em ECCV}, pages 499--515, 2018.

\bibitem{johnson2016densecap}
Justin Johnson, Andrej Karpathy, and Li~Fei-Fei.
\newblock Densecap: Fully convolutional localization networks for dense
  captioning.
\newblock In {\em CVPR}, pages 4565--4574, 2016.

\bibitem{kalos2009monte}
Malvin~H Kalos and Paula~A Whitlock.
\newblock {\em Monte carlo methods}.
\newblock John Wiley \& Sons, 2009.

\bibitem{karpathy2015deep}
Andrej Karpathy and Li~Fei-Fei.
\newblock Deep visual-semantic alignments for generating image descriptions.
\newblock In {\em CVPR}, pages 3128--3137, 2015.

\bibitem{kim2019dense}
Dong-Jin Kim, Jinsoo Choi, Tae-Hyun Oh, and In~So Kweon.
\newblock Dense relational captioning: Triple-stream networks for
  relationship-based captioning.
\newblock In {\em CVPR}, pages 6271--6280, 2019.

\bibitem{kingma2014adam}
Diederik~P Kingma and Jimmy Ba.
\newblock Adam: A method for stochastic optimization.
\newblock {\em arXiv preprint arXiv:1412.6980}, 2014.

\bibitem{krause2017hierarchical}
Jonathan Krause, Justin Johnson, Ranjay Krishna, and Li~Fei-Fei.
\newblock A hierarchical approach for generating descriptive image paragraphs.
\newblock In {\em CVPR}, pages 317--325, 2017.

\bibitem{krishna2017visual}
Ranjay Krishna, Yuke Zhu, Oliver Groth, Justin Johnson, Kenji Hata, Joshua
  Kravitz, Stephanie Chen, Yannis Kalantidis, Li-Jia Li, David~A Shamma, et~al.
\newblock Visual genome: Connecting language and vision using crowdsourced
  dense image annotations.
\newblock {\em IJCV}, 123(1):32--73, 2017.

\bibitem{kulkarni2013babytalk}
Girish Kulkarni, Visruth Premraj, Vicente Ordonez, Sagnik Dhar, Siming Li,
  Yejin Choi, Alexander~C Berg, and Tamara~L Berg.
\newblock Babytalk: Understanding and generating simple image descriptions.
\newblock {\em PAMI}, 35(12):2891--2903, 2013.

\bibitem{lecun1995convolutional}
Yann LeCun, Yoshua Bengio, et~al.
\newblock Convolutional networks for images, speech, and time series.
\newblock {\em The handbook of brain theory and neural networks},
  3361(10):1995, 1995.

\bibitem{li2020dual}
Ruifan Li, Haoyu Liang, Yihui Shi, Fangxiang Feng, and Xiaojie Wang.
\newblock Dual-cnn: A convolutional language decoder for paragraph image
  captioning.
\newblock {\em Neurocomputing}, 396:92--101, 2020.

\bibitem{liang2017recurrent}
Xiaodan Liang, Zhiting Hu, Hao Zhang, Chuang Gan, and Eric~P Xing.
\newblock Recurrent topic-transition gan for visual paragraph generation.
\newblock In {\em ICCV}, pages 3362--3371, 2017.

\bibitem{lin2014microsoft}
Tsung-Yi Lin, Michael Maire, Serge Belongie, James Hays, Pietro Perona, Deva
  Ramanan, Piotr Doll{\'a}r, and C~Lawrence Zitnick.
\newblock Microsoft coco: Common objects in context.
\newblock In {\em ECCV}, pages 740--755. Springer, 2014.

\bibitem{lu2017knowing}
Jiasen Lu, Caiming Xiong, Devi Parikh, and Richard Socher.
\newblock Knowing when to look: Adaptive attention via a visual sentinel for
  image captioning.
\newblock In {\em CVPR}, pages 375--383, 2017.

\bibitem{lu2022data}
Yue Lu, Chao Guo, Xingyuan Dai, and Fei-Yue Wang.
\newblock Data-efficient image captioning of fine art paintings via
  virtual-real semantic alignment training.
\newblock {\em Neurocomputing}, 2022.

\bibitem{mao2018show}
Yuzhao Mao, Chang Zhou, Xiaojie Wang, and Ruifan Li.
\newblock Show and tell more: Topic-oriented multi-sentence image captioning.
\newblock In {\em IJCAI}, pages 4258--4264, 2018.

\bibitem{melas2018training}
Luke Melas-Kyriazi, Alexander~M Rush, and George Han.
\newblock Training for diversity in image paragraph captioning.
\newblock In {\em EMNLP}, pages 757--761, 2018.

\bibitem{papineni2002bleu}
Kishore Papineni, Salim Roukos, Todd Ward, and Wei-Jing Zhu.
\newblock Bleu: a method for automatic evaluation of machine translation.
\newblock In {\em ACL}, pages 311--318. Association for Computational
  Linguistics, 2002.

\bibitem{peebles2020hessian}
William Peebles, John Peebles, Jun-Yan Zhu, Alexei Efros, and Antonio Torralba.
\newblock The hessian penalty: A weak prior for unsupervised disentanglement.
\newblock In {\em ECCV}, pages 581--597. Springer, 2020.

\bibitem{ren2015faster}
Shaoqing Ren, Kaiming He, Ross Girshick, and Jian Sun.
\newblock Faster r-cnn: Towards real-time object detection with region proposal
  networks.
\newblock In {\em Advances in neural information processing systems}, pages
  91--99, 2015.

\bibitem{rennie2017self}
Steven~J Rennie, Etienne Marcheret, Youssef Mroueh, Jerret Ross, and Vaibhava
  Goel.
\newblock Self-critical sequence training for image captioning.
\newblock In {\em CVPR}, pages 7008--7024, 2017.

\bibitem{tan2022acort}
Jia~Huei Tan, Ying~Hua Tan, Chee~Seng Chan, and Joon~Huang Chuah.
\newblock Acort: A compact object relation transformer for parameter efficient
  image captioning.
\newblock {\em Neurocomputing}, 2022.

\bibitem{vedantam2015cider}
Ramakrishna Vedantam, C~Lawrence~Zitnick, and Devi Parikh.
\newblock Cider: Consensus-based image description evaluation.
\newblock In {\em CVPR}, pages 4566--4575, 2015.

\bibitem{vinyals2015show}
Oriol Vinyals, Alexander Toshev, Samy Bengio, and Dumitru Erhan.
\newblock Show and tell: A neural image caption generator.
\newblock In {\em CVPR}, pages 3156--3164, 2015.

\bibitem{williams1992simple}
Ronald~J Williams.
\newblock Simple statistical gradient-following algorithms for connectionist
  reinforcement learning.
\newblock {\em Machine learning}, 8(3-4):229--256, 1992.

\bibitem{wu2020fine}
Jie Wu, Tianshui Chen, Hefeng Wu, Zhi Yang, Guangchun Luo, and Liang Lin.
\newblock Fine-grained image captioning with global-local discriminative
  objective.
\newblock {\em IEEE Transactions on Multimedia}, 23:2413--2427, 2021.

\bibitem{wu2019recall}
Lingxiang Wu, Min Xu, Jinqiao Wang, and Stuart Perry.
\newblock Recall what you see continually using gridlstm in image captioning.
\newblock {\em IEEE Transactions on Multimedia}, 22(3):808--818, 2020.

\bibitem{wu2019densely}
Siying Wu, Zheng-Jun Zha, Zilei Wang, Houqiang Li, and Feng Wu.
\newblock Densely supervised hierarchical policy-value network for image
  paragraph generation.
\newblock In {\em IJCAI}, pages 975--981. AAAI Press, 2019.

\bibitem{xu2015show}
Kelvin Xu, Jimmy Ba, Ryan Kiros, Kyunghyun Cho, Aaron Courville, Ruslan
  Salakhudinov, Rich Zemel, and Yoshua Bengio.
\newblock Show, attend and tell: Neural image caption generation with visual
  attention.
\newblock In {\em ICML}, pages 2048--2057, 2015.

\bibitem{yan2020paracnn}
Shiyang Yan, Yang Hua, and Neil Robertson.
\newblock Paracnn: Visual paragraph generation via adversarial twin contextual
  cnns.
\newblock {\em arXiv preprint arXiv:2004.10258}, 2020.

\bibitem{yang2017dense}
Linjie Yang, Kevin Tang, Jianchao Yang, and Li-Jia Li.
\newblock Dense captioning with joint inference and visual context.
\newblock In {\em CVPR}, pages 2193--2202, 2017.

\bibitem{yang2018multitask}
Min Yang, Wei Zhao, Wei Xu, Yabing Feng, Zhou Zhao, Xiaojun Chen, and Kai Lei.
\newblock Multitask learning for cross-domain image captioning.
\newblock {\em IEEE Transactions on Multimedia}, 21(4):1047--1061, 2019.

\bibitem{yao2018exploring}
Ting Yao, Yingwei Pan, Yehao Li, and Tao Mei.
\newblock Exploring visual relationship for image captioning.
\newblock In {\em ECCV}, pages 684--699, 2018.

\bibitem{yao2017boosting}
Ting Yao, Yingwei Pan, Yehao Li, Zhaofan Qiu, and Tao Mei.
\newblock Boosting image captioning with attributes.
\newblock In {\em ICCV}, pages 4894--4902, 2017.

\bibitem{yin2019context}
Guojun Yin, Lu~Sheng, Bin Liu, Nenghai Yu, Xiaogang Wang, and Jing Shao.
\newblock Context and attribute grounded dense captioning.
\newblock In {\em CVPR}, pages 6241--6250, 2019.

\end{thebibliography}

{\appendix
\section{Deduction of the disentanglement}
We add no extra constraint on $h_{e,i}^{(j)}$ and $h_{y,i}^{(j)}$ to reduce their correlation since the employed single addition layer under word-level cross-entropy loss facilitates the disentanglement property.

For simplicity, we reuse the symbol $\mathbf{x}\in \mathbb{R}^D$ and $\mathbf{y}\in \mathbb{R}^D$ to denote the semantic element $h_{e,i}^{(j)}$ and the syntactic element $h_{y,i}^{(j)}$, respectively. The input vector to the \textit{softmax} layer is $\mathbf{z}$ and the output is $\mathbf{f}$. Thus we have,
\begin{align}
    \mathbf{z}=\mathbf{W}(\mathbf{x}+\mathbf{y})~~\mathrm{and}~~
    \mathbf{f}=\sigma(\mathbf{z}),
\end{align}
where $\mathbf{W}$ is the parameter matrix and $\sigma(\cdot)$ is \textit{softmax} function.

Given that the $i$-th ground-truth word in the $j$-th sentence is an one-hot word vector $\mathbf{f}^*$ with $f_k=1$, the cross-entropy loss is
\begin{align}
    \mathcal{\ell}=-\log f_k,
\end{align}
and we have that
\begin{align}
    \frac{\partial f_k}{\partial \mathbf{x}}&=\frac{\partial f_k}{\partial \mathbf{z}}\cdot\frac{\partial \mathbf{z}}{\partial \mathbf{x}}=\frac{\partial f_k}{\partial \mathbf{z}}\cdot \mathbf{W}, \\
    \frac{\partial f_k}{\partial \mathbf{y}}&=\frac{\partial f_k}{\partial \mathbf{z}}\cdot\frac{\partial \mathbf{z}}{\partial \mathbf{y}}=\frac{\partial f_k}{\partial \mathbf{z}}\cdot \mathbf{W}.
\end{align}
Then according to gradient chain rule, we can derive 
\begin{align}
    \frac{\partial \mathcal{\ell}}{\partial \mathbf{x}}=-\frac{\partial \log f_k}{\partial \mathbf{x}}=-\frac{1}{f_k}\frac{\partial f_k}{\partial \mathbf{x}}.
\end{align}

To analyze the correlation between $\mathbf{x}$ and $\mathbf{y}$, we calculate the second-order partial derivatives as~\cite{peebles2020hessian} does
\begin{align} 
    \frac{\partial^2 \mathcal{\ell}}{\partial \mathbf{x} \partial \mathbf{y}}
    &=\frac{\partial}{\partial \mathbf{y}}\frac{\partial \mathcal{\ell}}{\partial \mathbf{x}}
    =-\frac{\partial}{\partial \mathbf{y}}\left(\frac{1}{f_k}\frac{\partial f_k}{\partial \mathbf{x}}\right) \label{eq:l_second} \\
    &=\frac{1}{f_k^2}\frac{\partial f_k}{\partial \mathbf{y}}\frac{\partial f_k}{\partial \mathbf{x}} - \frac{1}{f_k}\frac{\partial^2  f_k}{\partial\mathbf{x}\partial\mathbf{y}}. \label{eq:l_second2}
\end{align}
The two terms in Eq.~(\ref{eq:l_second2}) regardless of coefficient are
\begin{align} \label{eq:term1}
    \frac{\partial f_k}{\partial \mathbf{y}}\frac{\partial f_k}{\partial \mathbf{x}}
    &=\left(\frac{\partial f_k}{\partial x_t}\frac{\partial f_k}{\partial y_s}\right) \\ \nonumber
    &=\left(\left(\sum_i \frac{\partial f_k}{\partial z_i}w_{it}\right) \left(\sum_j \frac{\partial f_k}{\partial z_j}w_{js}\right)\right) \\ \nonumber
    &=\left(\sum_{i,j} \frac{\partial f_k}{\partial z_i}\frac{\partial f_k}{\partial z_j}w_{it}w_{js} \right),
\end{align}
and
\begin{align} \label{eq:f_second}
    \frac{\partial^2 f_k}{\partial\mathbf{x}\partial\mathbf{y}}
    &=\frac{\partial}{\partial \mathbf{y}}\left(\frac{\partial f_k}{\partial \mathbf{x}}\right)=\frac{\partial}{\partial \mathbf{y}}\left(\frac{\partial f_k}{\partial \mathbf{z}}\cdot \mathbf{W}\right)\\ \nonumber
    &=\left(\frac{\partial}{\partial y_s}\frac{\partial f_k}{\partial x_t}\right) =\left(\frac{\partial}{\partial y_s}\left(\sum_i \frac{\partial f_k}{\partial z_i}w_{it}\right) \right) \\ \nonumber
    &=\left(\sum_{i,j} \frac{\partial^2 f_k}{\partial z_i\partial z_j}\frac{\partial z_j}{\partial y_s} w_{it} \right) \\ \nonumber
    &=\left(\sum_{i,j}\frac{\partial^2 f_k}{\partial z_i \partial z_j}w_{js}w_{it}\right)
    =\mathbf{W}^T\left(\frac{\partial^2 f_k}{\partial \mathbf{z}^2}\right) \mathbf{W}.
\end{align}
Replacing Eq.~(\ref{eq:l_second}) with Eq.~(\ref{eq:term1}) and Eq.~(\ref{eq:f_second}), we have that
\begin{align} \label{eq:l_mat}
    \frac{\partial^2 \mathcal{\ell}}{\partial\mathbf{x}\partial\mathbf{y}}
    &=\frac{1}{f_k^2}\left(\sum_{i,j}w_{it}w_{js}\left(\frac{\partial f_k}{\partial z_i}\frac{\partial f_k}{\partial z_j}-f_k \frac{\partial^2 f_k}{\partial z_i \partial z_j} \right)\right) \\ \nonumber
    &{\doteq}\frac{1}{f_k^2}\mathbf{W}^T \mathbf{A} \mathbf{W},
\end{align}
where $\mathbf{A}=(a_{ij})=\left(\frac{\partial f_k}{\partial z_i}\frac{\partial f_k}{\partial z_j}-f_k \frac{\partial^2 f_k}{\partial z_i \partial z_j} \right)$.

Since $\mathbf{f}=\sigma(\mathbf{z})$, we have $f_k=\frac{e^{z_k}}{\sum_m e^{z_m}}$. The partial derivative of the \textit{softmax} function gives
\begin{align} \label{eq:softmaxt_first}
    \frac{\partial f_k}{\partial z_i}=\left\{\begin{array}{ll}
        -f_i f_k & \mathrm{if}~~~ i\neq k  \\
        f_k(1-f_k) & \mathrm{if}~~~ i=k.
    \end{array} \right.
\end{align}
Based on Eq.~(\ref{eq:softmaxt_first}), the second-order partial derivative is
\begin{align}
    \frac{\partial^2 f_k}{\partial z_i \partial z_j}{=}\left\{\begin{array}{ll}
        f_k(2f_k-1)(f_k{-}1) & \mathrm{if}\quad i=j=k \\
        f_j f_k(2f_k{-}1) & \mathrm{if}\quad i=k\neq j \\
        f_i f_k (2f_k{-}1) & \mathrm{if}\quad i\neq j=k \\
        f_k f_i (2f_i {-}1) & \mathrm{if}\quad i=j\neq k \\
        2f_i f_j f_k & \mathrm{if}\quad i\neq j \neq k.
    \end{array} \right.
\end{align}
Correspondingly, the elements of $\mathbf{A}$ can be calculated as
\begin{align} \label{eq:aij}
    a_{ij}=\left\{\begin{array}{ll}
        f_k^3(1-f_k) & \mathrm{if}\quad i=j=k~(\textrm{\textcircled{1}})\\
        -f_k^3 f_j & \mathrm{if}\quad i=k\neq j~(\textrm{\textcircled{2}}) \\
        -f_i f_k^3 & \mathrm{if}\quad i\neq j=k~(\textrm{\textcircled{3}}) \\
        f_k^2f_i(1-f_i) & \mathrm{if}\quad i=j\neq k~ (\textrm{\textcircled{4}}) \\
        -f_i f_j f_k^2 & \mathrm{if}\quad i\neq j\neq k~(\textrm{\textcircled{5}}).
    \end{array} \right.
\end{align}
Supposing $k=1$, a visualization of $\mathbf{A}$ is


\begin{align*}
    \mathbf{A}&=\left[\begin{array}{ccccc}
a_{11}& \tempa & a_{12}&\cdots &a_{1n}\\ \cline{1-5}
a_{21}& \tempa & a_{22}&\cdots &a_{2n}\\
\vdots& \tempa & \vdots&\ddots &\vdots \\
a_{n1}& \tempa &a_{n2}&\cdots &a_{nn}\\
\end{array}\right] \doteq\left[\begin{array}{c|c}
\textrm{\textcircled{1}} & \textrm{\textcircled{2}} \\ \cline{1-2}
\textrm{\textcircled{3}} & \begin{array}{ccc}
\textrm{\textcircled{4}} & & \textrm{\textcircled{5}}\\
& \ddots & \\
\textrm{\textcircled{5}} & & \textrm{\textcircled{4}} \\
\end{array} \end{array}\right].
\end{align*}

Minimizing the loss $\mathcal{\ell}$ leads to $f_k\to 1$ and $f_m \to 0$ for $m\neq k$. Thus, $a_{ij}\to 0$ (by Eq.~(\ref{eq:aij})) and $\mathbf{A}\to \mathbf{0}$. We assume that the parameter matrix $\mathbf{W}$ is bounded (\ie, $w_{ij}$ is finite), then it derives $\frac{\partial^2  \mathcal{\ell}}{\partial\mathbf{x}\partial\mathbf{y}}\to \mathbf{0}$ and $\frac{\partial^2  \mathcal{\ell}}{\partial\mathbf{y}\partial\mathbf{x}}\to \mathbf{0}$ by Eq.~(\ref{eq:l_mat}). 

Thus, minimizing the loss $\mathcal{\ell}$ also minimizes the off-diagonal entries of the function’s Hessian matrix \textit{w.r.t.} its inputs $\mathbf{x}$ and $\mathbf{y}$, which encourages the disentanglement between the inputs. Therefore, no extra constraint is compulsory to reduce the correlation of $h_{e,i}^{(j)}$ and $h_{y,i}^{(j)}$.

\section{More qualitative examples}
\begin{figure*}[!ht]
    \centering
    \includegraphics[width=\textwidth]{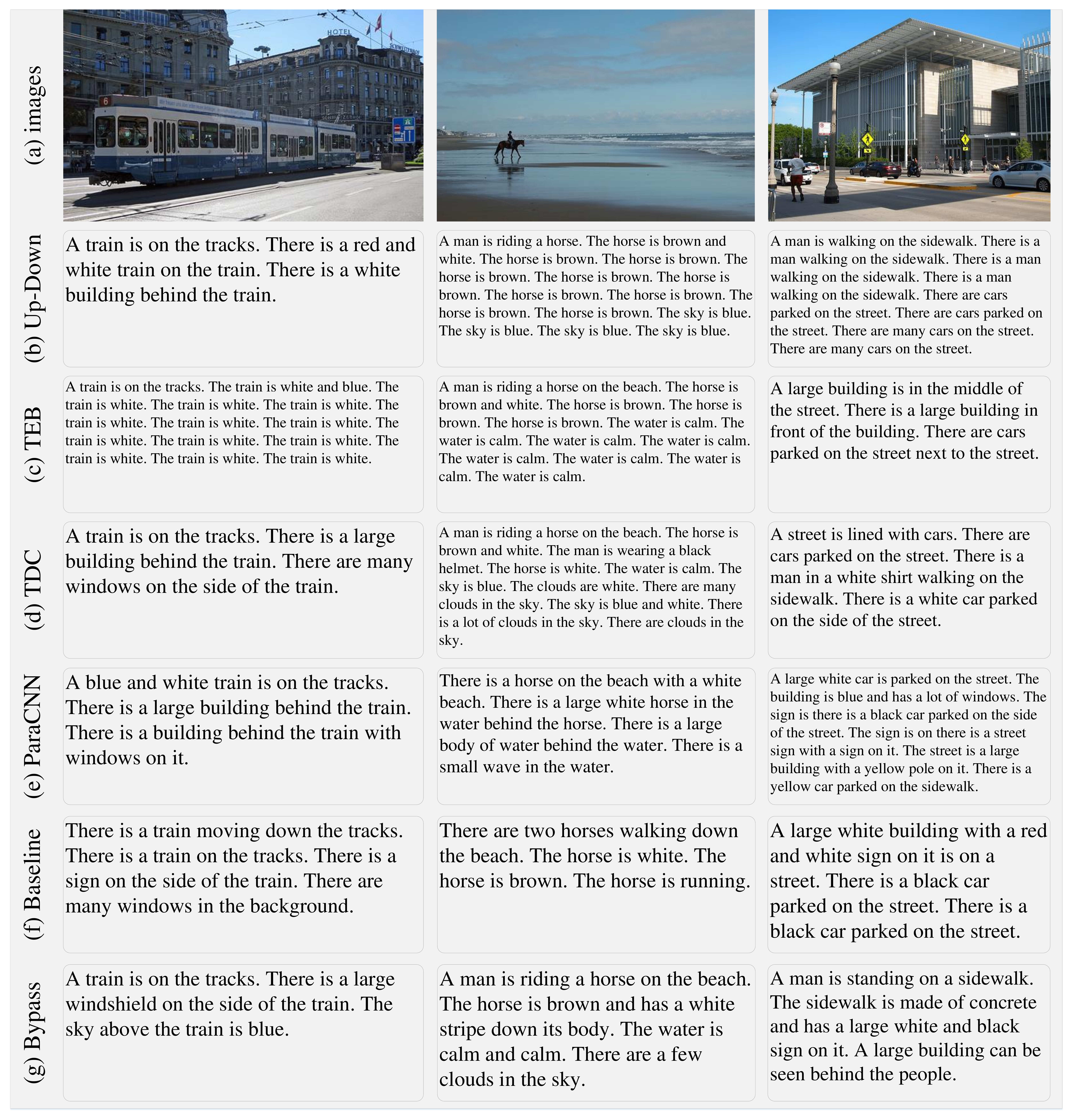}
    \caption{Descriptive captioning paragraphs generated by Up-Down, TDC, TEB, ParaCNN, the Baseline, and the Bypass model.
    }
    \label{fig:quality_supp}
\end{figure*}
Fig.~\ref{fig:quality_supp} is a more complete version of the quality comparison. We can see that `big' sentence methods Up-Down and TEB suffer from both immediate repetition and delayed repetition. TDC reduces immediate repetition by blocking trigrams, while ParaCNN and the Baseline model address it by explicitly modeling topic transition. However, they still suffer from delayed repetition. Our Bypass reduces both types of repetition and enhances the coherence of generated descriptive paragraphs.
}

\end{document}